\documentclass[]{bytedance_seed}

\usepackage{amsmath}
\usepackage{amssymb}
\usepackage{mathtools}
\usepackage{amsthm}
\usepackage{nicefrac}
\usepackage{algorithm}
\usepackage{algorithmic}
\usepackage{framed}
\usepackage{svg}
\usepackage{wrapfig}
\usepackage[export]{adjustbox}

\usepackage{amsmath,amsfonts,bm}

\def\eqref#1{equation~\ref{#1}}

\def\1{\bm{1}}

\DeclareMathAlphabet{\mathsfit}{\encodingdefault}{\sfdefault}{m}{sl}
\SetMathAlphabet{\mathsfit}{bold}{\encodingdefault}{\sfdefault}{bx}{n}

\title{Improved Large Language Diffusion Models}

\author[1,2,3,\dagger]{Shen Nie}
\author[4]{Qiyang Min}
\author[1,2,3]{Shaoxuan Xu}
\author[4]{Zihao Huang}
\author[4]{Yuxuan Song}
\author[4]{Yong Shan}
\author[1,2,3]{Yankai Lin}
\author[1,2,3]{Wayne Xin Zhao}
\author[1,2,3,*]{Chongxuan Li}
\author[1,2,3,*]{Ji-Rong Wen}

\affiliation[1]{Gaoling School of Artificial Intelligence, Renmin University of China}
\affiliation[2]{Beijing Key Laboratory of Research on Large Models and Intelligent Governance}
\affiliation[3]{Engineering Research Center of Next-Generation Intelligent Search and Recommendation, MOE}
\affiliation[4]{ByteDance Seed}

\contribution[\dagger]{Work done during an internship at ByteDance Seed}
\contribution[*]{Corresponding authors}

\checkdata[Contact]{\texttt{\{nieshen,chongxuanli\}@ruc.edu.cn}}
\checkdata[Correspondence]{Chongxuan Li and Ji-Rong Wen}

\abstract{
Modern large language models are predominantly trained with autoregressive factorization and causal attention. We present \emph{iLLaDA}, an 8B masked diffusion language model trained from scratch with fully bidirectional attention. iLLaDA keeps the masked diffusion objective throughout pre-training and supervised fine-tuning (SFT), scaling pre-training to 12T tokens and fine-tuning on a 25B-token instruction corpus for 12 epochs. We further use variable-length generation for efficiency and introduce confidence-based scoring for multiple-choice evaluation. Compared with LLaDA, iLLaDA improves broadly across general, mathematical, and code benchmarks; for example, iLLaDA-Base improves by 21.6 points on BBH and 14.9 points on ARC-Challenge, while iLLaDA-Instruct improves by 14.5 points on MATH and 16.5 points on HumanEval. Despite its non-autoregressive training, iLLaDA also remains competitive with Qwen2.5 7B on several benchmarks. These results show that fully bidirectional diffusion training from scratch is a competitive path toward strong language models. Model weights and codes: \url{https://github.com/ML-GSAI/LLaDA}.
}

\begin{document}

\maketitle

\section{Introduction}
\label{sec:introduction}

Large language models (LLMs) are currently dominated by the autoregressive paradigm~\citep{zhao2023survey,brown2020language,chatgpt,dubey2024llama}. Recently, diffusion language models have attracted increasing attention as a different approach to language generation. Following the masked diffusion formulation~\citep{austin2021structured,lou2023discrete,shi2024simplified,sahoo2024simple,ou2024your}, LLaDA trains a language model from scratch with fully bidirectional attention~\citep{nie2026large}. It shows that a non-autoregressive model can acquire core LLM capabilities such as in-context learning and instruction-following, challenging the common belief that language intelligence must rely on autoregressive modeling.

Beyond this conceptual implication, bidirectional diffusion language models have shown advantages in reversal and bidirectional reasoning~\citep{nie2026large,nie2024scaling}, long-horizon planning~\citep{ye2025beyondplanning}, and multimodal or omni-modeling~\citep{you2025llada,yang2025mmada,you2026lladao,xin2025luminadimoo}. Recent studies further show that bidirectional diffusion pre-training can better exploit limited data under repeated training, enabling diffusion language models to outperform autoregressive models in data-constrained settings~\citep{prabhudesai2025diffusion,ni2025diffusion}. However, LLaDA was still an initial large-scale attempt, and its performance remained behind strong autoregressive models such as Qwen2~\citep{qwen2} and Qwen2.5~\citep{qwen2.5}, leaving substantial room to improve.

We introduce \emph{iLLaDA} (\emph{improved LLaDA}), an 8B fully bidirectional masked diffusion language model trained from scratch. For pre-training, iLLaDA scales the corpus to 12T tokens, uses grouped-query attention~\citep{ainslie2023gqa} to reduce cache-style inference memory and tied input/output embeddings to reduce parameter count, and modifies the learning-rate schedule for large-scale training. For post-training, iLLaDA modifies the SFT strategy for variable-length generation and trains on a 25B-token instruction corpus for 12 epochs. For inference and evaluation, iLLaDA uses variable-length generation for efficiency and confidence-based scoring for multiple-choice benchmarks.

Experiments show that these changes substantially improve LLaDA. Compared with previous bidirectional diffusion language models, including LLaDA trained from scratch and Dream fine-tuned from Qwen2.5~\citep{ye2025dream}, iLLaDA obtains the best average performance in both base and instruction-tuned evaluations. Against Qwen2.5 7B~\citep{qwen2.5}, iLLaDA-Base is slightly stronger on average, while iLLaDA-Instruct still lags behind Qwen2.5 7B Instruct. Ablations further show that confidence-based scoring improves multiple-choice evaluation and that iLLaDA continues to benefit from SFT over multiple epochs.

\section{Approach}
\label{sec:approach}

This section describes the training and inference procedures of iLLaDA. We keep the masked diffusion formulation of LLaDA~\citep{austin2021structured,lou2023discrete,shi2024simplified,sahoo2024simple,ou2024your,nie2026large}, while making several practical changes that are important for scaling, post-training, and evaluation.

\subsection{Pre-training}
\label{sec:pretraining}

iLLaDA follows the same pre-training objective as LLaDA. Given a clean sequence \(x_0\) of length \(L\), we sample a masking ratio \(t \sim U[0, 1]\), independently replace each token by the mask token \(\textrm{M}\) with probability \(t\), and obtain a corrupted sequence \(x_t\). The model is trained to predict all masked tokens:
\begin{align}
\label{eq:pretrain-objective}
   \mathcal{L}(\theta)
   \triangleq
   -\mathbb{E}_{t, x_0, x_t}
   \left[
   \frac{1}{t}
   \sum_{i=1}^{L}
   \textbf{1}[x_t^i = \textrm{M}]
   \log p_{\theta}(x_0^i \mid x_t)
   \right].
\end{align}
This objective is a likelihood-based masked diffusion objective for discrete data~\citep{shi2024simplified,sahoo2024simple,ou2024your}, where the indicator function $\textbf{1}[\cdot]$ ensures that the loss is computed only for masked tokens. It differs from masked language modeling with a fixed masking ratio~\citep{devlin2018bert}.

The backbone of iLLaDA is a dense Transformer, which uses RMSNorm~\citep{zhang2019root}, SwiGLU~\citep{shazeer2020glu}, RoPE~\citep{su2024roformer}, and no attention or MLP bias. In contrast to LLaDA, which uses multi-head attention, iLLaDA uses grouped-query attention (GQA)~\citep{ainslie2023gqa}. Recent work has shown that KV-cache-like mechanisms can be adapted to diffusion language models~\citep{ma2025dkvcache,nguyentri2025attention,cheong2026entropycache,yang2025diffusionvar,qian2026d3llm}; under such cache-style implementations, GQA reduces the memory footprint of cached key/value states. To further control the parameter count, iLLaDA ties the input embedding and LM-head parameters. The architectural differences between iLLaDA and LLaDA are summarized in Tab.~\ref{tab:architecture}.

We pre-train iLLaDA with maximum sequence length 8192. We randomly split an 8192-token sequence into two shorter segments with probability \(30\%\), inspired by random-length training for masked diffusion language models~\citep{nie2024scaling}. We pack variable-length examples in each batch and compute attention with a FlashAttention-based variable-length attention kernel\footnote{\url{https://docs.pytorch.org/docs/2.12/nn.attention.varlen.html}}, which uses cumulative sequence offsets to separate examples without padding them to a common length. The learning rate is linearly warmed up to \(2 \times 10^{-4}\) and then kept constant. During training, when we observed that the pretraining loss stopped decreasing, we switched to a cosine decay schedule with minimum learning rate \(5 \times 10^{-6}\), after which the pretraining loss continued to improve. We use the AdamW optimizer~\citep{loshchilov2017decoupled} with weight decay 0.1.

\begin{table}[t!]
    \centering
    \caption{\textbf{Architecture comparison between iLLaDA and LLaDA.}}
    \label{tab:architecture}
    \vspace{.2cm}
    \begin{adjustbox}{max width=\textwidth}
    \begin{tabular}{lcc}
      \toprule
        & iLLaDA 8B & LLaDA 8B \\
      \midrule
      Layers & 32 & 32 \\
      Model dimension & 4096 & 4096 \\
      Attention heads & 32 & 32 \\
      Key/Value heads & 8 & 32 \\
      FFN dimension & 14,336 & 12,288 \\
      Vocabulary size & 155,136 & 126,464 \\
      Maximum sequence length & 8192 & 4096 \\
      Embedding and LM-head & Tied & Untied \\
      Total parameters & 7.62B & 8.02B \\
      Non-embedding parameters & 6.98B & 6.98B \\
      \bottomrule
    \end{tabular}
    \end{adjustbox}
    \vspace{-.2cm}
\end{table}

\subsection{Supervised Fine-Tuning}
\label{sec:sft}

Prior works~\citep{nie2026large,zhu2025llada,you2025llada} typically construct each SFT instance by concatenating a prompt with its full reference response. During training, the prompt tokens are kept visible, while masks are applied only within the response region. Within each mini-batch, shorter responses are padded with \( |\text{EOS}| \) tokens to match the length of the longest response.

iLLaDA instead uses the same data processing and masking scheme as pre-training. We format each instruction example as a prompt-response sequence followed by a single terminal \( |\text{EOS}| \), concatenate all formatted examples into a continuous instruction corpus, and sample 8192-token training sequences from this corpus. We then apply random masks to the entire sequence and optimize Eq.~(\ref{eq:pretrain-objective}), so prompt tokens, response tokens, and \( |\text{EOS}| \) tokens may all be masked. We also use the same random-length training as in pre-training. This SFT format naturally supports the variable-length block generation described in Sec.~\ref{sec:inference}.

Our SFT corpus contains approximately 25 billion tokens, and we fine-tune for 12 epochs. As shown in Sec.~\ref{sec:ablation}, our ablation study shows that iLLaDA continues to improve as the number of SFT epochs increases. During SFT, the learning rate is first linearly warmed up to \(5 \times 10^{-6}\), then kept constant, and finally linearly decayed to \(5 \times 10^{-7}\) over the last \(10\%\) of training. We use the AdamW optimizer~\citep{loshchilov2017decoupled} with weight decay 0.1.

\subsection{Inference}
\label{sec:inference}

iLLaDA uses the same probabilistic formulation as LLaDA~\citep{nie2026large}. In particular, both models are trained with the masked diffusion objective in Eq.~(\ref{eq:pretrain-objective}), which corresponds to an upper bound on the negative log-likelihood of the
model distribution.

Many language-model benchmarks are formulated as multiple-choice tasks, such as HellaSwag~\citep{zellers2019hellaswag}, PIQA~\citep{bisk2020piqa}, and ARC-Challenge~\citep{clark2018think}. Given a prefix \(p\) and a finite set of candidate continuations, evaluation requires assigning a score to each candidate and selecting the highest-scoring one. For iLLaDA, we use a deterministic confidence-based scoring rule, which performs better empirically than the upper-bound of log-likelihood on multiple-choice tasks.

Given a candidate continuation \(y\) of length \(L\), we start from an all-masked candidate and repeatedly reveal one ground-truth candidate token. At step \(k\), among the remaining masked positions \(\mathcal{M}_{k-1}\), we choose the token that the model assigns the highest confidence to:
\begin{align}
\label{eq:confidence-score}
i_k
=
\arg\max_{i \in \mathcal{M}_{k-1}}
p_{\theta}(y^i \mid p, \tilde{y}_{k-1}),
\quad
S_{\mathrm{conf}}(y \mid p)
=
\sum_{k=1}^{L}
\log p_{\theta}(y^{i_k} \mid p, \tilde{y}_{k-1}),
\end{align}
where \(\tilde{y}_{k-1}\) contains the revealed ground-truth tokens and masks elsewhere. This confidence score is not a likelihood estimate; rather, it is a task-specific scoring surrogate for comparing a finite set of candidate answers.

For open-ended generation, iLLaDA uses variable-length generation. Given a prompt, we append a block of mask tokens and run the diffusion sampler within this block. At each sampling step, the model predicts all masked positions, and we transfer the most confident predictions to visible tokens while keeping low-confidence positions masked, following the low-confidence remasking strategy of LLaDA and MaskGIT~\citep{chang2022maskgit}. Once a block is decoded, generation terminates if an \( |\text{EOS}| \) or other stop token appears; otherwise, a new block of masks is appended and the process continues until a maximum generation budget is reached.

\section{Experiments}
\label{sec:experiments}

In this section, we evaluate the base and instruction-following capabilities of iLLaDA on standard benchmarks, followed by ablation studies on multiple-choice scoring and SFT duration. The results show that iLLaDA substantially improves over prior diffusion language models and remains competitive with strong autoregressive baselines on several reasoning benchmarks.

\begin{table*}[t!]
    \centering
    \caption{\textbf{Benchmark Results of Base Models.} Results marked by $^{\dagger}$ and $^{\ddagger}$ are from~\citet{nie2026large} and~\citet{ye2025dream}, respectively. For Dream, 18T denotes Qwen2.5 pre-training tokens and 0.6T denotes diffusion fine-tuning tokens.}
    \label{tab:base}
    \begin{adjustbox}{max width=\textwidth}
    \begin{tabular}{l|cccc}
      \toprule
          & iLLaDA 8B & LLaDA 8B$^{\dagger}$ & Dream 7B$^{\ddagger}$ & Qwen2.5 7B$^{\ddagger}$ \\
      \midrule
      Model & Diffusion & Diffusion & Diffusion & AR \\
      Training tokens & 12T & 2.3T & 18T + 0.6T & 18T \\
      \midrule
      \multicolumn{5}{c}{General Tasks}\\
      \midrule
      MMLU & \textbf{74.8} & 65.9 & 69.5 & 71.9 \\
      BBH & \textbf{71.3} & 49.7 & 57.9 & 63.9 \\
      ARC-C & \textbf{60.8} & 45.9 & 59.8 & 51.5 \\
      Hellaswag & 76.6 & 70.5 & 73.3 & \textbf{79.0} \\
      \midrule
      \multicolumn{5}{c}{Mathematics \& Science}\\
      \midrule
      GSM8K & \textbf{81.9} & 70.3 & 77.2 & 78.9 \\
      Math & 38.4 & 31.4 & 39.6 & \textbf{41.1} \\
      \midrule
      \multicolumn{5}{c}{Code} \\
      \midrule
      HumanEval & 50.0 & 35.4 & \textbf{57.9} & 56.7 \\
      MBPP & 57.8 & 40.0 & 56.2 & \textbf{63.6} \\
      \midrule
      Average & \textbf{63.9} & 51.1 & 61.4 & 63.3 \\
      \bottomrule
    \end{tabular}
    \end{adjustbox}
\end{table*}

\begin{table*}[t!]
    \centering
    \caption{\textbf{Benchmark Results of Instruct Models.} Results marked by $^{\dagger}$ and $^{\ddagger}$ are from~\citet{nie2026large} and~\citet{ye2025dream}, respectively.}
    \label{tab:sft}
    \begin{adjustbox}{max width=\textwidth}
    \begin{tabular}{l|cccc}
      \toprule
          & iLLaDA 8B & LLaDA 8B$^{\dagger}$ & Dream 7B$^{\ddagger}$ & Qwen2.5 7B$^{\ddagger}$ \\
      \midrule
      Model & Diffusion & Diffusion & Diffusion & AR \\
      \midrule
      \multicolumn{5}{c}{General Tasks}\\
      \midrule
      MMLU & 71.6 & 65.5 & 67.0 & \textbf{76.6} \\
      MMLU-Pro & 52.3 & 37.0 & 43.3 & \textbf{56.3} \\
      MMLU-Redux & \textbf{76.4} & 68.9 & 76.3 & 75.7 \\
      \midrule
      \multicolumn{5}{c}{Mathematics \& Science}\\
      \midrule
      GSM8K & 89.0 & 77.5 & 81.0 & \textbf{91.6} \\
      Math & 56.7 & 42.2 & 39.2 & \textbf{75.5} \\
      \midrule
      \multicolumn{5}{c}{Code} \\
      \midrule
      HumanEval & 65.9 & 49.4 & 55.5 & \textbf{84.8} \\
      MBPP & 58.0 & 41.0 & 58.8 & \textbf{79.2} \\
      \midrule
      Avg. & 67.1 & 54.5 & 60.2 & \textbf{77.1} \\
      \bottomrule
    \end{tabular}
    \end{adjustbox}
    \vspace{-.2cm}
\end{table*}

\subsection{Benchmark Results}
\label{sec:results}

We evaluate iLLaDA in both base and instruction-tuned settings. The benchmark suite covers general language understanding and reasoning, including MMLU~\citep{hendrycks2020measuring}, BBH~\citep{suzgun2022challenging}, ARC-Challenge~\citep{clark2018think}, and HellaSwag~\citep{zellers2019hellaswag}; mathematical reasoning, including GSM8K~\citep{cobbe2021training} and MATH~\citep{hendrycks2021measuring}; and code generation, including HumanEval~\citep{chen2021evaluating} and MBPP~\citep{austin2021program}. For instruction-tuned models, we additionally report MMLU-Pro~\citep{wang2024mmlupro}, a more challenging multi-task understanding benchmark, and MMLU-Redux~\citep{gema2024done}, an error-corrected re-annotation of MMLU. We compare with representative diffusion language models, LLaDA 8B~\citep{nie2026large} and Dream 7B~\citep{ye2025dream}, as well as the autoregressive Qwen2.5 7B~\citep{qwen2.5}.

Tab.~\ref{tab:base} compares base models. iLLaDA substantially improves over LLaDA across all tasks, with particularly large gains on BBH, ARC-Challenge, GSM8K, HumanEval, and MBPP. Compared with Dream 7B, iLLaDA achieves stronger results on most general and mathematical benchmarks, while Dream remains stronger on HumanEval. Against Qwen2.5 7B, iLLaDA is competitive despite using a diffusion formulation, and obtains the best results on MMLU, BBH, ARC-Challenge, and GSM8K among the models reported in the table.

Tab.~\ref{tab:sft} reports instruction-tuned results. iLLaDA continues to outperform LLaDA and Dream on most benchmarks after SFT, and the improvements are especially pronounced on GSM8K, MATH, and HumanEval. Compared with Qwen2.5 7B, iLLaDA remains behind on several math and code benchmarks, but achieves competitive results on MMLU-Redux and substantially narrows the gap between diffusion language models and strong autoregressive baselines. Since iLLaDA Base is already competitive with Qwen2.5 Base in Tab.~\ref{tab:base}, we believe the remaining gap in the instruct setting is largely due to the additional reinforcement-learning alignment used by Qwen2.5 after SFT. We leave reinforcement-learning alignment for iLLaDA to future work. Please refer to Appendix~\ref{app:evaluation} for evaluation details.

\subsection{Ablation Studies}
\label{sec:ablation}

We first ablate the scoring rule for multiple-choice evaluation. As shown in Tab.~\ref{tab:ablation-llh}, confidence-based scoring consistently improves over the likelihood-style multiple-choice baseline, with gains of 1.3 on PIQA, 0.6 on ARC-Challenge, and 2.3 on HellaSwag. This result motivates the use of confidence-based scoring for the multiple-choice evaluations in Sec.~\ref{sec:inference}.

\begin{figure*}[t!]
  \vspace{0.15cm}
  \centering
  \begin{subfigure}{0.32\textwidth}
    \centering
    \includegraphics[width=\linewidth]{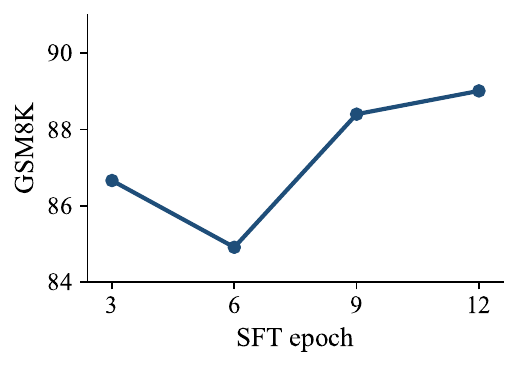}
  \end{subfigure}\hfill
  \begin{subfigure}{0.32\textwidth}
    \centering
    \includegraphics[width=\linewidth]{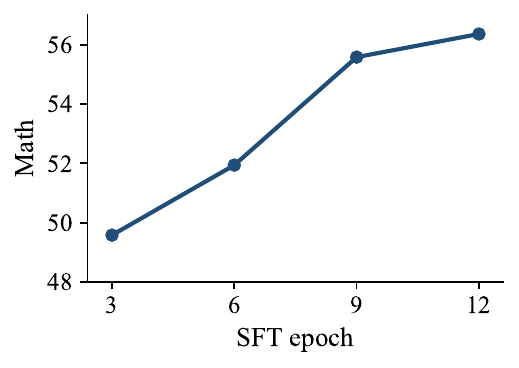}
  \end{subfigure}\hfill
  \begin{subfigure}{0.32\textwidth}
    \centering
    \includegraphics[width=\linewidth]{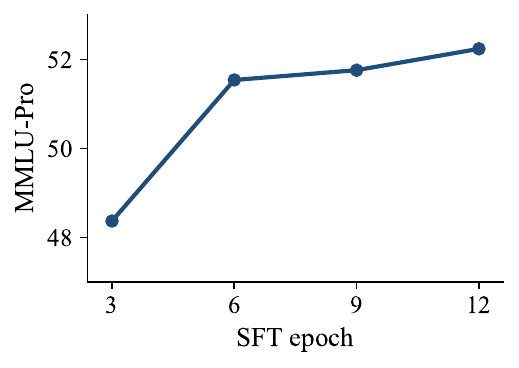}
  \end{subfigure}
  \caption{\textbf{SFT epoch ablation.} We evaluate iLLaDA at different SFT epochs on GSM8K, MATH, and MMLU-Pro.}
  \label{fig:sft-epochs}
  
\end{figure*}

We further study the effect of SFT duration. Fig.~\ref{fig:sft-epochs} shows that performance generally improves as the number of SFT epochs increases, supporting the use of long SFT for iLLaDA, especially on reasoning-heavy benchmarks. This observation is consistent with recent studies of diffusion language models in data-constrained regimes~\citep{prabhudesai2025diffusion,ni2025diffusion}; for example, \citet{ni2025diffusion} show that diffusion language models can continue to improve under extreme repeated pre-training settings, such as training on 1B unique tokens for 96 epochs. Our results suggest that a similar data-reuse effect also appears in SFT, where the instruction corpus is much smaller than the pre-training corpus and thus more relevant to practical instruction tuning. Due to compute constraints, we did not train beyond 12 SFT epochs.

\begin{table*}[t!]
    \centering
    \caption{\textbf{Ablation Results of Multiple-Choice Scoring Rules.}}
    \label{tab:ablation-llh}
    \begin{adjustbox}{max width=\textwidth}
    \begin{tabular}{l|ccc}
      \toprule
         Scoring rule & PIQA & ARC-C & Hellaswag \\
      \midrule
        Likelihood & 77.2 & 60.2 & 74.3 \\
        Confidence & \textbf{78.5} & \textbf{60.8} & \textbf{76.6} \\
      \bottomrule
    \end{tabular}
    \end{adjustbox}
\end{table*}

\section{Conclusion and Discussion}
\label{sec:conclusion}

We present iLLaDA, an 8B fully bidirectional diffusion language model trained from scratch. iLLaDA scales pre-training to 12T tokens and updates several parts of the practical recipe, including the model design, learning-rate schedule, SFT format, confidence-based multiple-choice scoring, and variable-length generation. We also find that iLLaDA continues to benefit from SFT over multiple epochs. Across base and instruction-tuned evaluations, these changes lead to substantial improvements over LLaDA on general, mathematical, and code benchmarks, suggesting that fully bidirectional diffusion training from scratch can achieve strong language modeling performance.

This report also leaves several limitations. First, iLLaDA has not been further aligned with reinforcement learning, which may partly explain the remaining gap between iLLaDA-Instruct and strong autoregressive instruct models. Recent RL methods developed for masked diffusion LLMs, such as VRPO, diffu-GRPO, MDPO, and ESPO~\citep{zhu2025llada,zhao2025d1,he2025mdpo,ou2025principled}, can be directly applied to iLLaDA and are likely to further improve its instruction-following and reasoning abilities. Second, due to limited compute, our study is limited to the 8B scale and does not provide a fully matched comparison with autoregressive models; instead, we allocate our compute to 12T-token pre-training. We leave reinforcement-learning alignment and larger-scale studies for future work.


\bibliographystyle{unsrtnat}
\bibliography{main}

\newpage
\appendix

\section{Evaluation Details}
\label{app:evaluation}

This appendix provides additional details for the evaluations in Sec.~\ref{sec:experiments}.

For iLLaDA-8B-Base, we use open-ended generation for BBH, GSM8K, MATH, HumanEval, and MBPP. For BBH, GSM8K, MATH, and MBPP, we set the maximum generation length to 1024 and the block length to 32. For HumanEval, we set both the maximum generation length and the block length to 512, since we observed that semi-autoregressive block sampling hurts performance on this benchmark.

For iLLaDA-8B-Instruct, we use benchmark-specific inference settings. For MMLU and MMLU-Redux, where the model only needs to generate a single answer letter, we set the maximum generation length and block length to 4/4 and 3/3, respectively. For GSM8K and HumanEval, we set the maximum generation length to 2048 and the block length to 32. For MMLU-Pro and MATH, we set the maximum generation length to 4096 and the block length to 32. For MBPP, we set the maximum generation length to 2048 and the block length to 16.

For iLLaDA-8B-Instruct, we observe repetitive reasoning loops on some difficult problems, where the model may repeatedly produce phrases such as ``Wait, let me check again'' and fail to produce a final answer. We attribute this behavior to a subset of the SFT corpus that contains structured chain-of-thought traces generated by reasoning models. To mitigate such loops, as generation becomes longer, we gradually increase the probability of emitting the stop-thinking token \texttt{\textless/think\textgreater}, encouraging the model to terminate the reasoning trace and produce the final answer.

\end{document}